# Ranking medical jargon in electronic health record notes by adapted distant supervision

Jinying Chen[1], Abhyuday N. Jagannatha[2], Samah J. Jarad[3], Hong Yu[4,1]

[1] Department of Quantitative Health Sciences, University of Massachusetts Medical School, Worcester, MA, USA
[2] School of Computer Science, University of Massachusetts, Amherst, MA, USA
[3] Yale Center for Medical Informatics, Yale University, New Haven, CT, USA
[4] Bedford Veterans Affairs Medical Center, Center for Healthcare Organization and Implementation Research, Bedford, MA, United States
{jinying.chen, hong.yu}@umassmed.edu, abhyuday@cs.umass.edu, samah.fodeh@yale.edu

## ABSTRACT

**Objective:** Allowing patients to access their own electronic health record (EHR) notes through online patient portals has the potential to improve patient-centered care. However, medical jargon, which abounds in EHR notes, has been shown to be a barrier for patient EHR comprehension. Existing knowledge bases that link medical jargon to lay terms or definitions play an important role in alleviating this problem but have low coverage of medical jargon in EHRs. We developed a data-driven approach that mines EHRs to identify and rank medical jargon based on its importance to patients, to support the building of EHR-centric lay language resources.

**Methods:** We developed an innovative adapted distant supervision (ADS) model based on support vector machines to rank medical jargon from EHRs. For distant supervision, we utilized the open-access, collaborative consumer health vocabulary, a large, publicly available resource that links lay terms to medical jargon. We explored both knowledge-based features from the Unified Medical Language System and distributed word representations (word embeddings) learned from unlabeled large corpora. We evaluated the ADS model using physician-identified important medical terms.

**Results:** Our ADS model significantly surpassed two state-of-the-art automatic term recognition methods, TF*IDF and C-Value, yielding 0.810 ROC-AUC versus 0.710 and 0.667, respectively. Our model identified over 10K important medical jargon terms after ranking over 100K candidate terms mined from over 7,500 EHR narratives.

**Conclusion:** Our work is an important step towards enriching lexical resources that link medical jargon to lay terms/definitions to support patient EHR comprehension. The identified medical jargon terms and their rankings are available upon request.

## INTRODUCTION

Patient portals, including My HealtheVet [1], have been embraced by many healthcare organizations for patient-clinician communication. Allowing patients to access their EHR notes helps improve their disease understanding, self-management and outcomes [1,2]. However, studies have shown that patients often have difficulty in comprehending medical jargon [3–7] (here "medical jargon" is defined as "technical terminology or special words that are used by medical professions and are difficult for others to understand"), limiting their ability to understand their clinical status [5,6]. Figure 1 shows a sample text found in a typical clinical note. The medical terms that may hinder patients' comprehension are italicized. In addition, those medical jargon terms judged by physicians to be important for patient understanding are also underlined.

**Figure 1**. Illustration of medical jargon in a clinical note

> Her *creatinine* has shown a steady rise over the past four years. She does have *nephritic range proteinuria*. The likely *etiology* of her *nephrotic range proteinuria* is her diabetes.
> She was on an *ACE inhibitor*, which was just stopped in August due to the elevated *creatinine* of 4.41. Given the severity of her *nephrotic syndrome*, her chronic kidney disease is likely permanent; however, I will repeat a *chem-8* now that she is off the *ACE inhibitor*. I will also get a *renal duplex scan* to make sure she does not have any *renal artery stenosis*.

To reduce the communication gap between patients and clinicians, there have been decades of research efforts in creating medical resource for lay people [8]. Natural language processing methods have also been developed to automatically substitute medical jargon with lay terms [9–11] or to link them to consumer-oriented definitions [12]. These approaches require high-quality lexical resources of medical jargon and lay terms/definitions. The open-access collaborative consumer health vocabulary (CHV) is one such resource [13]. It has been incorporated into the Unified Medical Language System and has also been used in EHR simplification [9,10].

Research in CHV has been motivated by the vocabulary discrepancies between lay people and health care professionals [14–17]. CHV incorporates terms extracted from various consumer health sites, such as queries submitted to MedLinePlus, a consumer-oriented online knowledge resource maintained by the National Library of Medicine, and from postings in health-focused online discussion forums [18,19]. CHV contained 152,338 terms, most of which are consumer health terms [18–20]. Zeng et al. [18] mapped these consumer health terms to the Unified Medical Language System by a semi-automatic approach. As the result of this work, CHV encompasses lay terms as well as corresponding medical jargon.

From our current work, we found that CHV alone is not sufficient as a lexical resource for comprehending EHR notes, as many medical jargon terms in EHRs do not exist in CHV, and many others exist in CHV but with lay terms identical to the jargon terms themselves. For example, in CHV, the respective lay terms of 19,674 jargon terms (e.g., "neurocytoma", "lymphangiomatosis", and "laryngeal carcinoma") are themselves. Although CHV provides lay definitions to some of these terms, 18,823 (96%) terms remain to be unannotated (i.e., have neither appropriate lay terms nor lay definitions).

The goal of this study is to identify medical jargon from EHRs for the purpose of creating new lexical entries to link medical jargon to lay terms/definitions. Since the size of medical jargon is large (tens of thousands of terms), we will rank them based on how important they are to lay people, and therefore prioritize the annotation effort of lexical entries on those important terms. Specifically, we proposed and developed an adapted distant-supervision (ADS) model to rank terms in EHRs to prioritize those important medical jargon terms for lay language annotation. We made a novel use of a non-EHR-centric resource (i.e., CHV) for distant supervision and showed promising results from using this approach. Our work is different from previous work in building biomedical lexical resources. Previous work either uses unsupervised automatic term recognition methods [21–23] or uses supervised learning (when human annotations are available) [21] to prioritize terms.

Our contributions are twofold. Firstly, we develop and evaluate the ADS model, a new model that mines and ranks medical terms from large EHR corpora based on terms' importance to patients. Secondly, we apply our ADS model to rank over 100K EHR terms to prioritize 10K important terms for lay language annotation.

In addition, the ranking methods we developed have a great potential to be applied to other clinical natural language processing tasks, including generating features for keyphrase extraction, information retrieval, summarization, and question answering.

## MATERIALS AND METHODS

### EHR Corpora and Candidate Terms

In this study, we utilized two EHR corpora: EHR-Pittsburg and EHR-UMass.

*EHR-Pittsburg* [1] contains 7,839 discharge summary notes with 5.4M words. We applied the linguistic filter of the automatic term recognition toolkit *Jate* [24] to this corpus and extracted 106,108 candidate terms (see Step 1 in Figure 2). These candidate terms were further used to identify and rank medical jargon terms.

**Figure 2**. Overview of our approach: data extraction (Steps 1 and 2), ADS (Step 3) and evaluation (Step 4)

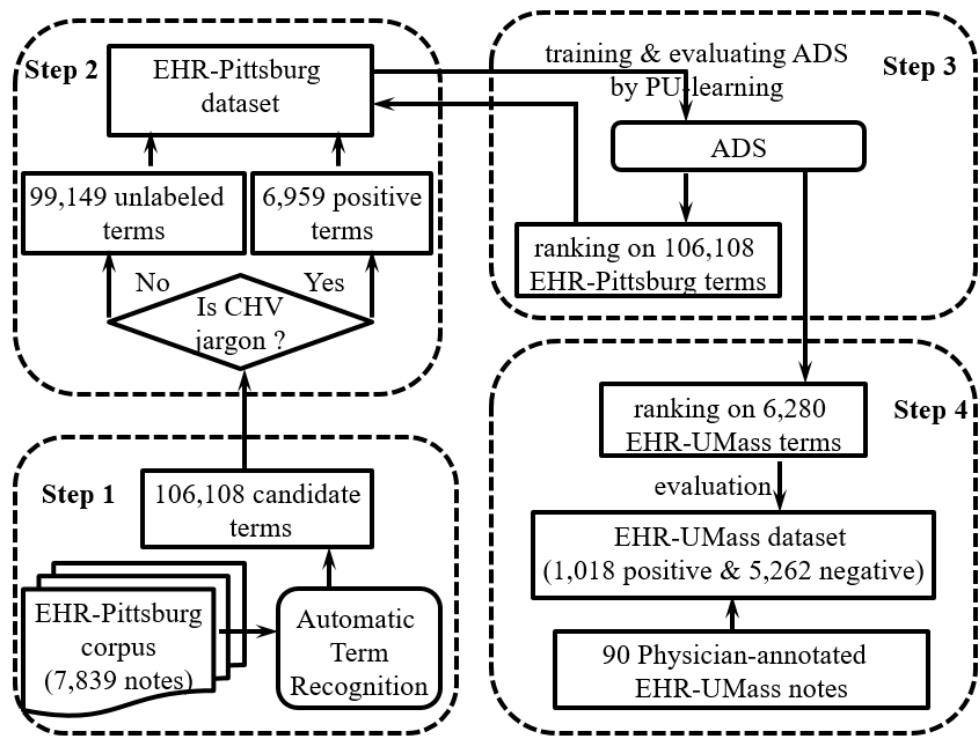

---

[1] Chapman W. University of Pittsburgh Natural Language Processing Repository (http://www.dbmi.pitt.edu/nlpfront). Using this data requires a license.

*EHR-UMass* contains 90 de-identified EHR notes from the UMass Memorial Hospital outpatient clinics. To maximize the representativeness, we selected notes from patients with six common primary clinical diagnoses: cancer, COPD, diabetes, heart failure, hypertension, and liver failure. We de-identified the notes and then asked physicians to identify, for each note, terms important to patients [25]. Specifically, physicians were asked to identify EHR terms which the patients need to know to comprehend the notes for the most important aspects medically relevant to their health and treatment course. For each note, we obtained annotations from two physicians. Three physicians did the annotation and annotated 48, 68, and 64 notes respectively. We used this expert-annotated corpus to create a dataset for evaluating ADS (details in the subsection Evaluation Using the EHR-UMass Dataset).

**Medical Jargon in CHV**

In this study, we evaluated the coverage of CHV for medical jargon in EHR. In addition, we used medical jargon in CHV to create training data for the ADS model. We followed [9] (i.e., CHV familiarity score ≤ 0.6) to identify medical jargon terms in CHV.

**Baseline Systems**

We evaluated two baseline systems for ranking medical terms: Corpus-level TF*IDF [24] and C-Value [26]. Both baselines are state-of-the-art automatic term recognition methods that have proved to be effective in identifying domain-specific biomedical terms [24,26]. EHR notes are abundant in medical jargon. The two methods, by their definitions described below, are expected to be able to identify these medical terms and rank them by their importance to the EHR corpora. When applying the two methods to our task, we made an assumption that medical terms salient in a large EHR corpus are important to patients (based on the fact that EHR notes are documents that record patients' medical and treatment course) and therefore should be prioritized for annotation.

*Corpus-level TF*IDF:* TF*IDF [27] is a widely used metric for measuring the importance of a term to a document *d* in a corpus *D*. The more frequently the term appears in the document and the less frequently it appears in other documents, the more important it is to this document. We used Corpus-level TF*IDF (abbreviated as TF*IDF in the rest of the paper) to measure the importance of a candidate term *t* to a corpus *D* by summing up a term's TF*IDF per document over the corpus, as defined in Equation (1):

$$TF*IDF(t,D) = \sum_{d \in D}(tf(t,d) \times idf(t,D)) \qquad (1)$$

where *d* is any document in *D*; *tf(t,d)* is term frequency of *t* in *d*; and *idf(t,D)* is inverse document frequency of *t* in *D*, which indicates the importance of *t* across the corpus *D* and is defined as $log\frac{|D|}{|\{d|t \in d\}|}$.

*C-Value:* C-Value is a widely used method for extracting terminology from text corpora. It has been used to prioritize health and biomedical terms for developing lexical resources, including CHV [22,23]. It measures the importance of a term *t* in a corpus *D* by its frequency in the corpus *tf(t,D)*. If a term is nested in other longer terms, C-Value penalizes it, as defined in Equation (2):

$$C-Value\ (t,D) = \begin{cases} log_2|t| \times tf(t,D), & a\ is\ not\ nested \\ log_2|t| \times tf(t,D) - \frac{1}{|T_t|}\sum_{b \in T_t} tf(b,D), & otherwise \end{cases} \qquad (2)$$

where *|t|* is the number of words contained in *t*; $T_t$ is the set of all long candidate terms (phrases) *b* that contain *t*; and *|$T_t$|* is the number of terms in $T_t$.

**The ADS Model**

Our ADS model is a case of learning from positive and unlabeled data [28–32]. In particular, Elkan and Noto [32] have shown that a binary classifier that outputs probability confidence scores on examples can be trained using positive examples and unlabeled examples (treated as "negative") and its confidence scores on new examples can then be used to rank those new examples.

We used CHV to select and label positive examples to train ADS (see Step 2 in Figure 2). Our approach is based on an assumption that medical terms important to patients must be used by patients. Specifically, we assume that medical terms that occur in both EHRs and CHV are important to patients because they are medical synonyms of terms initially identified from queries and postings generated by patients in online health forums. Based on this assumption, we used medical jargon terms in both EHRs and CHV (called EHR-CHV medical jargon terms) as positive examples because our goal is to prioritize important medical jargon terms from EHRs. In addition to the positive examples labeled by CHV, some unlabeled EHR terms may be also important and therefore should also be ranked high. We achieved this goal by learning from positive and unlabeled data. Figure 2 illustrates data extraction, ADS and its evaluation.

Our ADS model has two major components: the support vector machine classifier and the features used for classification.

The Support Vector Machine Classifier

Previous work has shown that support vector machines [33] are effective in learning from positive and unlabeled data [29,32,34], our ADS is therefore built upon support vector machines. We employed the widely used and reproducible LibSVM package [35]. We chose the RBF-kernel support vector machine as we found it performed better than support vector machines using linear and polynomial kernels in our preliminary experiments. We assigned a rank score to each term using LibSVM's probabilistic outputs [36,37]. We used these probabilistic rank scores to merge the rankings from 10-fold runs to obtain the global rank of a term.

Features

We developed three types of learning features: (1) confidence scores as computed by automatic term recognition (ATR) (2) Unified Medical Language System semantic types (Sem), and (3) distributed word representation or word embedding (WE).

*Confidence scores from automatic term recognition*: we used the confidence scores from TF*IDF and C-value.

*Unified Medical Language System semantic types:* We mapped candidate terms to Unified Medical Language System concepts and included semantic types for those concepts that had an exact match or a head-noun match as features. Each semantic type is a 0-1 binary feature. This type of feature has been used to identify domain-specific medical terms [12,22]. In this work, we made an assumption that different semantic types contribute differently to a term's importance to patients. We relied on the support vector machine classifier to learn the weight/contribution of each semantic type.

*Distributed word representation (word embedding)*: Word embeddings are distributed vector representations of words. Each dimension of a word vector has a real value ranged between 0 and 1. We treated each dimension as a feature.

Because word embedding vectors are learned from large text corpora and incorporate syntactic and semantic properties of words, words sharing similar semantics and context are expected to be close in their word vector space [38,39]. In this work, we made an assumption that medical terms that are important to patients share similar semantics and context. Therefore, word embeddings are likely to be useful features for learning important medical terms.

We trained a neural language model to learn word embeddings. Specifically, we used Word2Vec software, which supports efficient computations on large datasets, to create the Skip Gram word embeddings [38,39]. We trained Word2Vec using a combined text corpus (over 3G words) of English Wikipedia, articles from PubMed Open Access and 99K EHR notes from EHR-Pittsburg. We set the training parameters based on the study of Pyysalo et al. [40]. We represented multi-word terms with the mean of individual word vectors.

**Training and Evaluation Datasets**

We used two datasets in our study. The EHR-Pittsburg dataset was used for training, evaluation, and generating the global ranking of candidate terms, while the EHR-UMass dataset was used for evaluation. The statistics of these two datasets are summarized in Table 1.

| Dataset | EHR notes | # of terms | Positive | Unlabeled/negative | Purpose of use |
|---|---|---|---|---|---|
| EHR-Pittsburg | 7,839 | 106,108 | 6,959 | 99,149 (unlabeled) | Train and evaluate ADS, generate the global ranking of candidate terms |
| EHR-UMass | 90 | 6,280 | 1,018 | 5,262 (negative) | Evaluate ADS |

**Table 1**. A summary of the two datasets used in this study

Training Using the EHR-Pittsburg Dataset

The numbers of terms used as positive and unlabeled data were 6,959 and 99,149, respectively (see Step 2 in Figure 2). For training, we divided the data into 10 folds. We used 9 folds to train the ADS model and applied it to the remaining fold to obtain the rank scores of the candidate terms. We produced a total of 10 ranking outputs, one for each fold. We then merged the 10 outputs to produce a global ranking, which we evaluated using a metrics that measures system performance for learning from positive and unlabeled data (details in the subsection Evaluation Metrics).

Evaluation Using the EHR-UMass Dataset

Because the EHR-UMass corpus was annotated by physicians in such a way that terms that are important to patients were labeled as "positive", we utilized their annotations to create a data set with both positive and negative examples for evaluation (see Step 4 in Figure 2). Specifically, we applied *Jate* to extract 6,280 candidate terms from the EHR-UMass corpus. Candidate terms exactly matching the physician-annotated important terms were labeled as positive. The remaining candidate terms were labeled as negative. In total, we obtained 1,018 positive and 5,262 negative examples, which we included in the EHR-UMass dataset for evaluating ADS. We compute the C-Value and TF*IDF scores for the terms in this dataset by using a large EHR corpus that contains 6K notes (including the 90 EHR-UMass notes) collected using the same six diagnoses used for collecting the 90 notes.

**Post-processing**

As we found that the performances of TF*IDF and C-Value were negatively affected by high-frequency, common terms (e.g., "patient" and "pain") in EHRs, we added a post-processing procedure that used a stopwords list to filter out common terms from the models' outputs. This list contains 100 high-frequency non-medical terms in the EHR-UMass corpus. In addition, we used regular expressions to rank low compound terms that contain "not" "no", "and", or "or". In our evaluation using the EHR-UMass dataset, we report both conditions: with and without post-processing.

**Evaluation Metrics**

*Receiver Operating Characteristic (ROC) and Area Under ROC Curve (ROC-AUC):* ROC curve is a metrics widely used for evaluating ranking outputs. It plots the true positive rate (y-coordinate) against the false positive rate (x-coordinate) at various threshold settings. We report both ROC and ROC-AUC by using the R library pROC [41].

*Metrics for learning from Positive and Unlabeled data* (*PU Metrics*): Evaluating systems that learn from positive and unlabeled data is challenging because the data include unlabeled examples, and thus we can't calculate *recall* and *precision*. For evaluation, Lee and Liu [42] introduced *PU metrics*, $r^2/Pr[system\ positive]$, where *r is recall* and *Pr[system positive]* is the probability of positive examples predicted by the system. *Recall* can be estimated as the total number of positive predictions divided by the total number of labeled positive examples, as explained in [42]. We plotted $r^2/Pr[system\ positive]$ as a function of the rank *k*.

## RESULTS

**CHV Coverage of Medical Jargon in EHRs**

From the EHR-Pittsburg corpus, we extracted 106,108 candidate terms, on which we applied MetaMap [43] to select medical terms. A total of 19,503 (18%) of the candidate terms were successfully mapped to Unified Medical Language System concepts by MetaMap. However, 4,680 (24%) of these medical terms do not appear in CHV. We manually examined those terms and found a majority of them were medical jargon terms, such as "Bruton agammaglobulinemia", "molecular diagnostics", "motor symptom", and "reactive lymphocytosis". The remaining 86,605 (82%) candidate terms also contained medical jargon terms, such as "anticardiolipin", "BGM",[2] "demargination", "heptoglobin", "hypoalimentation", and "hypobilirubinemia".

**ADS Ranking Performance on EHR-Pittsburg Dataset**

Figure 3 plots the PU metrics as a function of rank *k* for the ADS model and two baseline systems on the EHR-Pittsburg dataset. The PU metrics curve of ADS rapidly reaches to the peak at *k* = 9,229, and then declines sharply. In contrast, the two baseline systems' PU metrics are relatively stable. Overall, ADS was consistently better than the two baselines for all *k*.

**ADS Ranking Performance on EHR-UMass Dataset**

Figure 4 plots the ROC curves of ADS, TF*IDF and C-Value in ranking EHR-UMass terms, without and with post-processing. ADS achieved the best performance. Table 2 shows the ROC-AUC, where ADS outperformed TF*IDF and C-Value by large margins (>15 points, absolute gains). Although post-processing improved performance of TF*IDF and C-Value substantially, ADS still exhibited better performance.

**Figure 3**. Plots of the PU metrics ($r^2/Pr[system\ positive]$) as a function of rank *k* for different methods in ranking the EHR-Pittsburg terms

---

[2] BGM ("blood glucose monitoring") is a frequently used acronym in clinical notes. It is only registered as a gene name in the Unified Medical Language System.

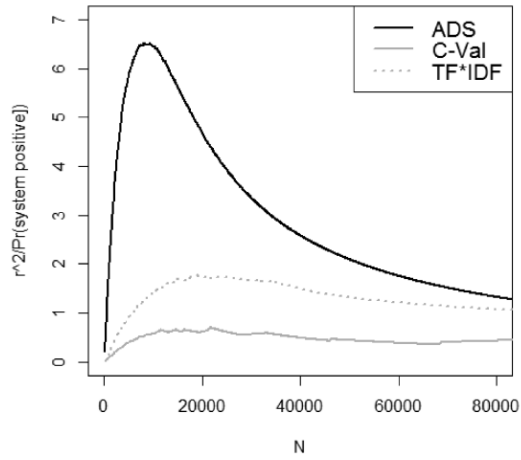

|  | TF*IDF | C-Value | ADS |
|---|---|---|---|
| ROC-AUC, without post-processing | 0.638 | 0. 556 | **0.788** |
| ROC-AUC, with post-processing | 0.710 | 0.666 | **0.810** |

**Table 2**. ROC-AUC values of different methods in ranking the EHR-UMass terms, without and with post-processing

**Figure 4**. ROC plots of different methods in ranking the EHR-UMass terms: (4a) without post-processing and (4b) with post-processing

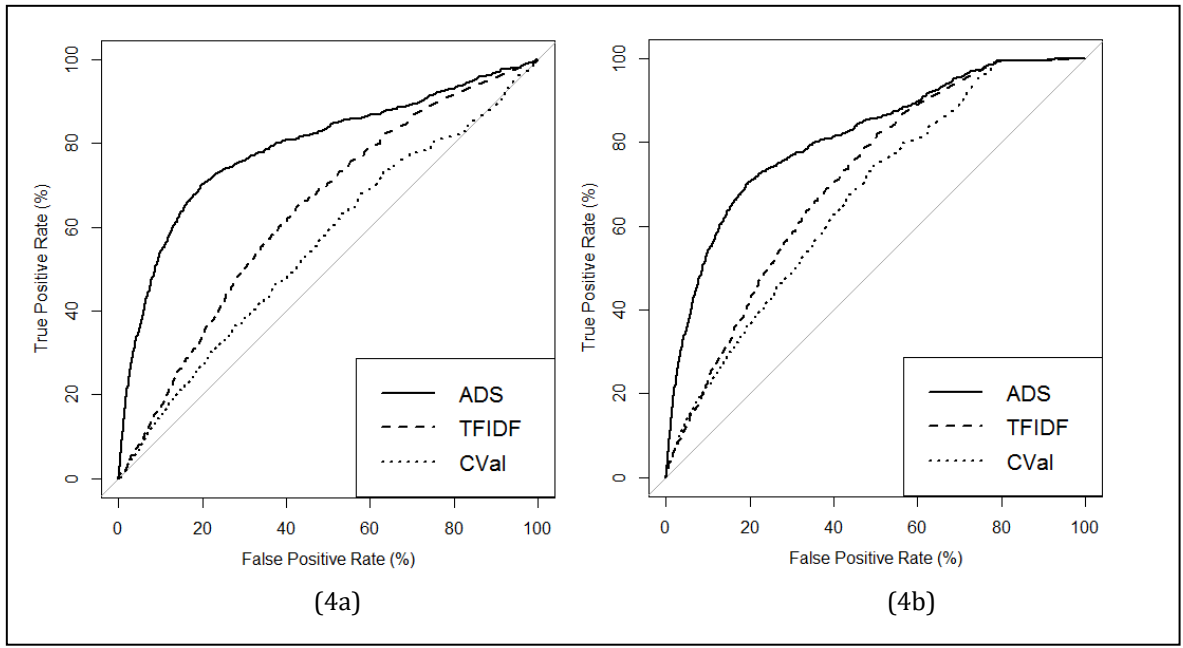

## DISCUSSION

**Principle Results**

We find that CHV has incomplete coverage of medical jargon in EHRs. We therefore developed the ADS model to rank 100K candidate terms from the EHR-Pittsburg corpus and prioritized our annotation of lay terms/definitions for top-ranked terms. ADS ranks EHR terms based on the assumption that medical terms that occur in both EHRs and CHV are important to patients. ADS achieved 0.810 ROC-AUC on the EHR-UMass dataset (Table 2). This level of performance is adequate, especially considering that ADS does not use any human-annotated training data. This result indirectly verifies the validity of our assumption. It also suggests that we can use ADS to prioritize EHR terms for annotation.

**Interpretation of the PU Metrics Curves**

In Figure 3, the performance of ADS on the EHR-Pittsburg dataset reaches a peak at rank $k = 9,229$, with a sharp drop afterwards. At its peak point, ADS is able to identify 5,248 (75%) of the total

6,959 labeled positive terms (i.e., EHR-CHV medical jargon terms). Most of them are important medical jargon, including "premature atrial contraction", "polycythemia", "T-cell lymphoma", and "thallium stress test". The non-CHV terms that were ranked by ADS in top-10K also contain many important medical jargon terms, such as "chronic respiratory insufficiency", "nasogastric decompression", and "preoperative chemotherapy". At lower ranks, ADS is less effective in identifying EHR-CHV jargon terms, finding more common terms such as "nose", "activate", "training", and "dust". This result suggests that we can use the rank *10K* as a threshold to divide the ranked terms into high-quality and low-quality groups. Terms in the high-quality group are being used to support annotation of lay terms/definitions.

The curves of TF*IDF and C-Value are relatively flat because term ranking is based on statistical values and therefore insensitive to CHV terms, as opposed to ADS, which is supervised by CHV.

**ADS and the Baselines TF*IDF and C-Value**

Figure 4 and Table 2 show that ADS outperforms the two baselines on the EHR-UMass dataset. Our result analysis shows that the three models have similar performance at top ranks (top-*n* lists where *n*<30) and ADS has much better performance at lower ranks. We manually checked the top-10 erroneous terms identified from the EHR-UMass dataset (with post-processing), shown in Table 4 where the EHR-CHV medical jargon terms are bolded. As shown in Table 4, the top-10 errors identified by ADS are all EHR-CHV medical jargon terms (bolded). Because the physicians only identified few (15 per note, on average) important medical terms from each EHR note, they did not mark many CHV terms as important. However, this type of error may not be critical for our annotation task. For example, some CHV terms not marked by physicians are still worth annotating with lay definitions/terms for EHR comprehension, e.g., "Raynaud"(Raynaud's disease), "pyelonephristis", "onychomycosis", and "cholestasis".

Although post-processing boosted TF*IDF and C-Value performances, they still ranked some common terms high (e.g., "surgery", "sleep", "liver", "abdominal pain", and "blood sugar"). In addition, C-Value ranks many multi-word terms high because it favors long phrases by its definition.

| ADS | **Raynaud, macrolide, polyuria, Tegretol, aminoglycoside, pyelonephritis, onychomycosis, cholestasis, coarctation, Imuran** |
|---|---|
| TF*IDF | surgery, lesion, area, sleep, continue, issue, liver, breast, Allscripts, state |
| C-Value | abdominal pain, past medical history, blood sugar, p.r.n., **normal limit**, CT scan, vital sign, family history, low back pain, soft nontender nondistended |

**Table 4**. Top-10 erroneous terms identified from the EHR-UMass dataset by ADS, TF*IDF and C-Value, with post-processing

The false-negative terms predicted by the three systems (i.e., important medical jargon terms that were ranked low) are also different. ADS often missed medical terms that contain easy words, such as "pancreatic enzyme replacement" and "chronic lower extremity edema". One reason for this error is that these terms have similar word embedding and/or semantic type features as lay (negative) terms (e.g., "replacement", "chronic", "lower", and "extremity"). Using advanced phrase embedding techniques may alleviate this problem, which we may explore in future. Different from ADS, TF*IDF and C-Value often missed terms that are important but occur infrequently in the 6K EHR-UMass notes, such as "neurodermatitis", "diabetic renal disease", "autonomic neuropathy", and "PML" (progressive multifocal leukoencephalopathy).

# CONCLUSION

We report an ADS model for prioritizing medical terms that are important for patient EHR comprehension. Our experiments have shown that ADS is more effective than TF*IDF and C-Value, two methods that are widely used to mine and prioritize terms from large text corpora for building domain-specific lexical resources. The EHR terms prioritized by our model are being used to enrich a comprehensive medical jargon–lay term/definition knowledge resource for EHR simplification. Our top-10K-ranked EHR terms are available upon request.

# ACKNOWLEDGMENTS

This work was supported by the Investigator Initiated Research (1I01HX001457-01) from the Health Services Research and Development Program of the United States Department of Veterans Affairs. The content is solely the responsibility of the authors and does not represent the views of the United States Department of Veterans Affairs or the United States Government.

We thank Weisong Liu for technical support in collecting the EHR notes and the UMassMed annotation team, including Elaine Freund, Victoria Wang, Andrew Hsu, Barinder Hansra and Sonali Harchandani, for creating the UMass-EHR corpus.

# REFERENCES

1 Nazi KM, Hogan TP, McInnes DK, *et al.* Evaluating patient access to Electronic Health Records: results from a survey of veterans. *Med Care* 2013;**51**:S52–S56.

2 Strum S. Inviting patients to read doctors' notes. *Ann Intern Med* 2012;**156**:608. doi:10.7326/0003-4819-156-8-201204170-00016

3 Lerner EB, Jehle DV, Janicke DM, *et al.* Medical communication: do our patients understand? *Am J Emerg Med* 2000;**18**:764–6. doi:10.1053/ajem.2000.18040

4   Chapman K, Abraham C, Jenkins V, *et al.* Lay understanding of terms used in cancer consultations. *Psychooncology* 2003;**12**:557–66. doi:10.1002/pon.673

5   Pyper C, Amery J, Watson M, *et al.* Patients' experiences when accessing their on-line electronic patient records in primary care. *Br J Gen Pr* 2004;**54**:38–43.

6   Keselman A, Slaughter L, Smith CA, *et al.* Towards consumer-friendly PHRs: patients' experience with reviewing their health records. In: *Proceedings of AMIA Annual Symposium*. 2007. 399–403.

7   Pieterse AH, Jager NA, Smets EMA, *et al.* Lay understanding of common medical terminology in oncology. *Psychooncology* 2013;**22**:1186–91. doi:10.1002/pon.3096

8   McCray AT. Promoting health literacy. *J Am Med Inform Assoc* 2005;**12**:152–163.

9   Zeng-Treitler Q, Goryachev S, Kim H, *et al.* Making texts in electronic health records comprehensible to consumers: a prototype translator. In: *Proceedings of AMIA Annual Symposium*. 2007. 846–50.

10  Kandula S, Curtis D, Zeng-Treitler Q. A semantic and syntactic text simplification tool for health content. In: *Proceedings of AMIA Annual Symposium*. 2010. 366–70.

11  Abrahamsson E, Forni T, Skeppstedt M, *et al.* Medical text simplification using synonym replacement: Adapting assessment of word difficulty to a compounding language. In: *Proceedings of the 3rd Workshop on Predicting and Improving Text Readability for Target Reader Populations at EACL.* Association for Computational Linguistics 2014. 57–65.

12  Polepalli Ramesh B, Houston T, Brandt C, *et al.* Improving Patients' Electronic Health Record Comprehension with NoteAid. *Stud Health Technol Inform* 2013;**192**:714–8.

13  Zeng QT, Tse T. Exploring and developing consumer health vocabularies. *J Am Med Inform Assoc* 2006;**13**:24.

14  McCray AT, Loane RF, Browne AC, *et al.* Terminology issues in user access to Web-based medical information. *Proc AMIA Symp* 1999;:107–11.

15  Zeng Q, Kogan S, Ash N, *et al.* Patient and clinician vocabulary: how different are they? *Stud Health Technol Inform* 2001;**84**:399–403.

16  Patrick TB, Monga HK, Sievert MC, *et al.* Evaluation of Controlled Vocabulary Resources for Development of a Consumer Entry Vocabulary for Diabetes. *J Med Internet Res* 2001;**3**:e24. doi:10.2196/jmir.3.3.e24

17  Zeng Q, Kogan S, Ash N, *et al.* Characteristics of consumer terminology for health information retrieval. *Methods Inf Med* 2002;**41**:289–298.

18    Zeng QT, Tse T, Crowell J, *et al.* Identifying consumer-friendly display (CFD) names for health concepts. In: *Proceedings of AMIA Annual Symposium*. 2005. 859–63.

19    Keselman A, Smith CA, Divita G, *et al.* Consumer health concepts that do not map to the UMLS: where do they fit? *J Am Med Inform Assoc JAMIA* 2008;**15**:496–505. doi:10.1197/jamia.M2599

20    Tse T, Soergel D. Exploring Medical Expressions Used by Consumers and the Media: An Emerging View of Consumer Health Vocabularies. *AMIA Annu Symp Proc* 2003;**2003**:674–8.

21    Zeng Q, Tse T, Divita G, *et al.* Term identification methods for consumer health vocabulary development. *J Med Internet Res* 2007;**9**:e4.

22    Spasić I, Schober D, Sansone S-A, *et al.* Facilitating the development of controlled vocabularies for metabolomics technologies with text mining. *BMC Bioinformatics* 2008;**9**:S5.

23    Doing-Harris KM, Zeng-Treitler Q. Computer-assisted update of a consumer health vocabulary through mining of social network data. *J Med Internet Res* 2011;**13**:e37. doi:10.2196/jmir.1636

24    Zhang Z, Iria J, Brewster C, *et al.* A Comparative Evaluation of Term Recognition Algorithms. In: *Proceedings of the sixth international conference of Language Resources and Evaluation (LREC)*. 2008. 2108–13.

25    Chen J, Zheng J, Yu H. Finding Important Terms for Patients in Their Electronic Health Records: A Learning-to-Rank Approach Using Expert Annotations. *JMIR Med Inform* (accept)

26    Frantzi K, Ananiadou S, Mima H. Automatic recognition of multi-word terms:. the c-value/nc-value method. *Int J Digit Libr* 2000;**3**:115–130.

27    Sparck Jones K. A statistical interpretation of term specificity and its application in retrieval. *J Doc* 1972;**28**:11–21.

28    Denis F, Gilleron R, Tommasi M. Text classification from positive and unlabeled examples. In: *Proceedings of the 9th International Conference on Information Processing and Management of Uncertainty in Knowledge-Based Systems, IPMU'02*. 2002. 1927–1934.

29    Liu B, Dai Y, Li X, *et al.* Building text classifiers using positive and unlabeled examples. In: *Data Mining, 2003. ICDM 2003. Third IEEE International Conference on*. IEEE 2003. 179–186.

30     Zhang D, Lee WS. A simple probabilistic approach to learning from positive and unlabeled examples. In: *Proceedings of the 5th Annual UK Workshop on Computational Intelligence (UKCI)*. Citeseer 2005. 83–87.

31     Fung GPC, Yu JX, Lu H, *et al.* Text classification without negative examples revisit. *Knowl Data Eng IEEE Trans On* 2006;**18**:6–20.

32     Elkan C, Noto K. Learning classifiers from only positive and unlabeled data. In: *Proceedings of the 14th ACM SIGKDD international conference on Knowledge discovery and data mining*. ACM 2008. 213–220.

33     Cortes C, Vapnik V. Support-vector networks. *Mach Learn* 1995;**20**:273–97. doi:10.1007/BF00994018

34     Mordelet F, Vert J-P. A bagging SVM to learn from positive and unlabeled examples. *Pattern Recognit Lett* 2014;**37**:201–209.

35     Chang C-C, Lin C-J. LIBSVM: a library for support vector machines. *ACM Trans Intell Syst Technol TIST* 2011;**2**:27:1-27:27.

36     Platt J. Probabilistic outputs for support vector machines and comparisons to regularized likelihood methods. *Adv Large Margin Classif* 1999;**10**:61–74.

37     Lin H-T, Lin C-J, Weng RC. A note on Platt's probabilistic outputs for support vector machines. *Mach Learn* 2007;**68**:267–276.

38     Mikolov T, Chen K, Corrado G, *et al.* Efficient Estimation of Word Representations in Vector Space. *ArXiv13013781 Cs* Published Online First: 16 January 2013.http://arxiv.org/abs/1301.3781

39     Mikolov T, Sutskever I, Chen K, *et al.* Distributed representations of words and phrases and their compositionality. In: *Advances in Neural Information Processing Systems*. 2013. 3111–9.

40     Pyysalo S, Ginter F, Moen H, *et al.* Distributional semantics resources for biomedical text processing. In: *The 5th International Symposium on Languages in Biology and Medicine*. 2013. 39–43.

41     Robin X, Turck N, Hainard A, *et al.* pROC: an open-source package for R and S+ to analyze and compare ROC curves. *BMC Bioinformatics* 2011;**12**:77.

42     Lee WS, Liu B. Learning with positive and unlabeled examples using weighted logistic regression. In: *ICML*. 2003. 448–455.

43     Aronson AR, Lang F-M. An overview of MetaMap: historical perspective and recent advances. *J Am Med Inform Assoc* 2010;**17**:229–36.